\documentclass[letterpaper]{article} 
\usepackage{aaai25}  
\usepackage{times}  
\usepackage{helvet}  
\usepackage{courier}  
\usepackage[hyphens]{url}  
\usepackage{graphicx} 
\usepackage{natbib}  
\usepackage{caption} 
\usepackage{algorithm}
\usepackage{algorithmic}
\usepackage{newfloat}
\usepackage{listings}
\DeclareCaptionStyle{ruled}{labelfont=normalfont,labelsep=colon,strut=off} 
\floatstyle{ruled}
\newfloat{listing}{tb}{lst}{}
\floatname{listing}{Listing}
\usepackage{amsmath,amssymb}
\usepackage{pifont}

\title{Image Classification Using a Diffusion Model as a Pre-Training Model}
\author {
    Kosuke Ukita\textsuperscript{\rm 1, \rm a},
    Ye Xiaolong\textsuperscript{\rm 1, \rm b},
    Tsuyoshi Okita\textsuperscript{\rm 1, \rm C},
}
\affiliations {
    \textsuperscript{\rm 1}Kyushu Institute of Technology\\
    \textsuperscript{\rm a}ukita.kosuke299@mail.kyutech.jp, 
    \textsuperscript{\rm b}ye.xiaolong713@mail.kyutech.jp, 
    \textsuperscript{\rm c}tsuyoshi@ai.kyutech.ac.jp
}

\usepackage{bibentry}

\begin{document}
\maketitle

\begin{abstract}
In this paper, we propose a diffusion model that integrates a representation-conditioning mechanism, where the representations derived from a Vision Transformer (ViT) are used to condition the internal process of a Transformer-based diffusion model. This approach enables representation-conditioned data generation, addressing the challenge of requiring large-scale labeled datasets by leveraging self-supervised learning on unlabeled data. We evaluate our method through a zero-shot classification task for hematoma detection in brain imaging. Compared to the strong contrastive learning baseline, DINOv2, our method achieves a notable improvement of +6.15\% in accuracy and +13.60\% in F1-score, demonstrating its effectiveness in image classification.
\end{abstract}

\section{Introduction} \label{sec:introduction}

Recent years have witnessed remarkable progress in image recognition powered by deep learning, with self-supervised learning emerging as a promising paradigm for learning high-quality feature representations from unlabeled data. Self-supervised learning typically involves defining pseudo-labels for unlabeled inputs and training models to predict them, thereby capturing the underlying structure of the data without requiring costly manual annotations. Among existing self-supervised learning approaches, contrastive learning has gained significant attention and achieved notable success through methods such as MoCo \cite{he2019moco}, SimCLR \cite{chen2020simple}, and DINO \cite{caron2021emerging}. These methods learn representations by encouraging consistency between different augmented views of the same image while distinguishing between views of different images.

Despite their effectiveness, contrastive methods have inherent limitations. Their focus on maximizing inter-instance discriminability tends to bias the model toward learning global features, often at the expense of local or fine-grained information. This makes them vulnerable to changes such as partial occlusion or viewpoint variation \cite{chuang2020debiased}. Furthermore, their reliance on carefully selected negative samples introduces challenges in stability and performance \cite{robinson2020contrastive}.

To address these issues, recent studies have explored diffusion models as an alternative for self-supervised representation learning. Compared to traditional generative models like GANs \cite{goodfellow2014generative} and VAEs \cite{kingma2013auto}, diffusion models enable more stable training and produce higher-quality images, making them particularly suitable for representation learning in image recognition. These models learn by gradually adding noise to the input data and then reconstructing the original image through a reverse denoising process. By explicitly modeling the data distribution, diffusion models are able to capture both global and local semantic structures in a consistent and robust manner. Furthermore, their resilience to viewpoint changes and partial occlusions has been demonstrated in prior work \cite{xiang2023denoising}, highlighting their potential to overcome the key limitations of contrastive learning.

While diffusion models have demonstrated potential for self-supervised representation learning, current research in this area remains relatively limited. Existing approaches often adapt diffusion models originally designed for generative tasks without explicitly optimizing them for downstream recognition performance. Furthermore, there is a lack of frameworks that integrate latent-level conditioning mechanisms to guide the denoising process in a representation-aware manner. This motivates the need for a new design that fully leverages the generative nature of diffusion while explicitly targeting effective feature extraction for classification.

In this work, we propose a novel self-supervised learning framework that leverages diffusion models for image representation learning. The core idea is to shift the focus from instance-level discrimination to a generative reconstruction-based paradigm, thereby learning rich and robust visual representations from unlabeled data. Specifically, we explore how the progressive denoising process in diffusion models can be utilized for pre-training in image classification tasks, offering an alternative and potentially more effective pathway for representation learning beyond contrastive approaches.

To this end, we introduce the Representation-Conditioned Latent Diffusion Transformer, a new architecture designed to enhance the effectiveness of self-supervised learning through diffusion. In this model, latent representations are used as conditioning signals within the diffusion process, allowing the model to effectively extract and refine image features during pretraining. The proposed approach aims to fully exploit the multi-scale representational capacity of diffusion models while maintaining training stability and flexibility across downstream tasks.

Our method introduces a novel conditioning mechanism where intermediate latent representations are injected into the denoising process via a Transformer-based architecture. This design allows for flexible integration of semantic information at different stages of the reverse process, improving the model’s ability to capture hierarchical features essential for recognition tasks.

The main contributions of this paper are summarized as follows:

\begin{itemize}
\item We propose a novel self-supervised learning framework based on diffusion models for visual representation learning.
\item We introduce the Representation-Conditioned Latent Diffusion Transformer, which leverages latent features to guide the denoising process and improve feature extraction quality.
\end{itemize}

Beyond image classification, the proposed framework provides a generalizable approach to pretraining that could benefit other vision tasks such as detection, segmentation, and multimodal learning, where robust feature representations from unlabeled data are crucial.

The structure of this paper is as follows: First, Chapter \ref{sec:relatedwork} describes related works. Chapter \ref{sec:propose} describes the proposed method. Next, Chapter \ref{sec:experiment} explains the experimental setup, and Chapter \ref{sec:results} shows the results. Finally, Chapter \ref{sec:conclusion} presents the conclusions of this research.

\section{Related Works} \label{sec:relatedwork}
Currently, DiT \cite{Peebles2022DiT} is one of the widely used diffusion models. DiT is a method based on the Latent Diffusion Model \cite{rombach2022highresolution} that trains the reverse diffusion process in latent space, and its denoising network is configured with an architecture based on the Transformer \cite{vaswani2023attention}. In this research, we adopt DiT as the base architecture. DiT allows for much faster processing compared to conventional U-Net-based diffusion models \cite{Peebles2022DiT}. Additionally, we believe it enables a fair comparison with models based on Vision Transformer (ViT) \cite{dosovitskiy2020image} (e.g., DINO \cite{caron2021emerging}), which are compared in this experiment. Therefore, this research validates the effectiveness of a DiT-based method.

\begin{figure}[h]
    \begin{center}
        \includegraphics[scale=0.48]{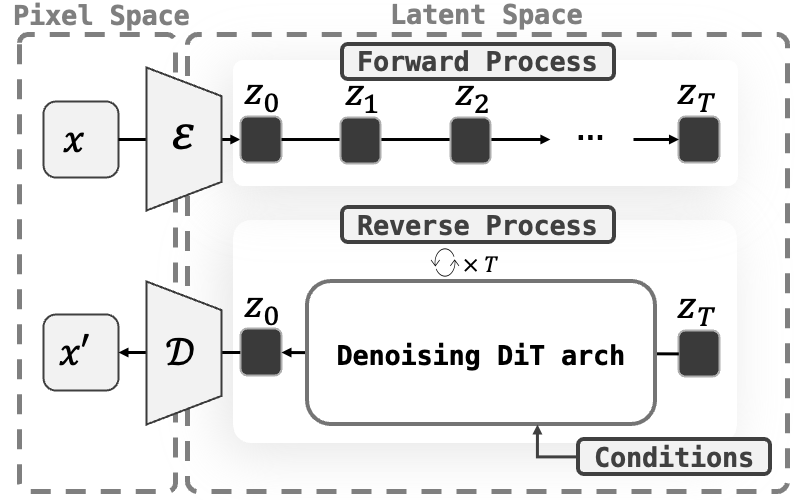}
        \caption{Latent Diffusion Transformer}
        \label{fig:dit}
    \end{center}
\end{figure}

Figure \ref{fig:dit} illustrates the model structure of DiT. DiT was proposed as a class-conditioned diffusion model, trained by providing class labels as conditions. In this paper, we propose a new approach for this conditioning, and thus we refer to the generalized model architecture for conditioning as a "Latent Diffusion Transformer." In other words, DiT can be regarded as a class-conditioned Latent Diffusion Transformer. Since conditioning on timesteps has been shown to improve performance \cite{ho2020denoising}, we assume that timesteps are included in the conditions. It should be noted that to focus on conditions other than timesteps, timesteps may be omitted in some descriptions in this paper.

Several studies have reported the use of diffusion models for image classification \cite{mukhopadhyay2023diffusion,xiang2023denoising,wei2023diffusion,yang2023diffusion}. In particular, Diffusion Classifier \cite{li2023diffusion} achieved zero-shot image classification utilizing diffusion models by using DiT. However, Diffusion Classifier presupposes the existence of a DiT trained on a large-scale labeled dataset, and the model learning and data annotation work for this incurs high costs. In this research, to overcome this challenge, we introduce a self-supervised learning framework that utilizes unlabeled datasets and propose a method to achieve high image classification performance with only a small amount of labeled data.

Research utilizing diffusion models in the context of self-supervised learning is still an emerging field, but several preceding studies have indicated its potential. Chen et al. \cite{chen2024deconstructing} analyzed the representation learning capability of diffusion models in detail and proposed a method to transform them into encoders as classical Denoising Autoencoders (DAE) \cite{vincent2008extracting} by incrementally decomposing the denoising process. Particularly, through experiments using DiT, they asserted that mapping to a latent space composed of PCA, VAEs, etc., is an important component in image classification tasks, demonstrating the utility of diffusion models in self-supervised learning.

\section{Proposed Methods} \label{sec:propose}
In Chapter \ref{sec:propose}, we detail a new method for treating diffusion models as a framework for self-supervised learning. This method, similar to approaches in conventional contrastive learning, is divided into two stages: pre-training with an unlabeled dataset and a downstream task with a small labeled dataset. To utilize the Latent Diffusion Transformer for self-supervised learning, we propose a Representation-Conditioned Latent Diffusion Transformer.

\subsection{Representation-Conditioned Latent Diffusion Transformer} \label{sec:dit_repcond}

\begin{figure}[h]
    \begin{center}
        \includegraphics[scale=0.53]{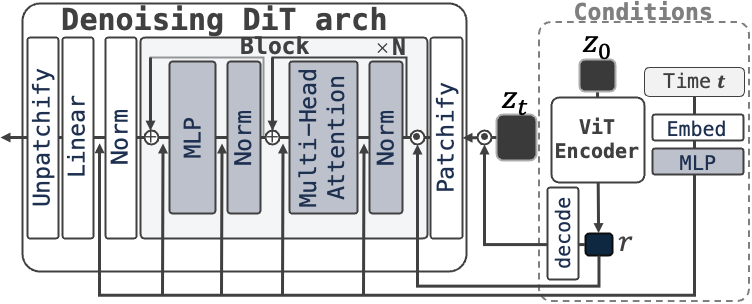}
        \caption{Representation-Conditioned Latent Diffusion Transformer}
        \label{fig:repcond_dit}
    \end{center}
\end{figure}
\begin{figure*}[h]
    \begin{center}
        \includegraphics[scale=0.7]{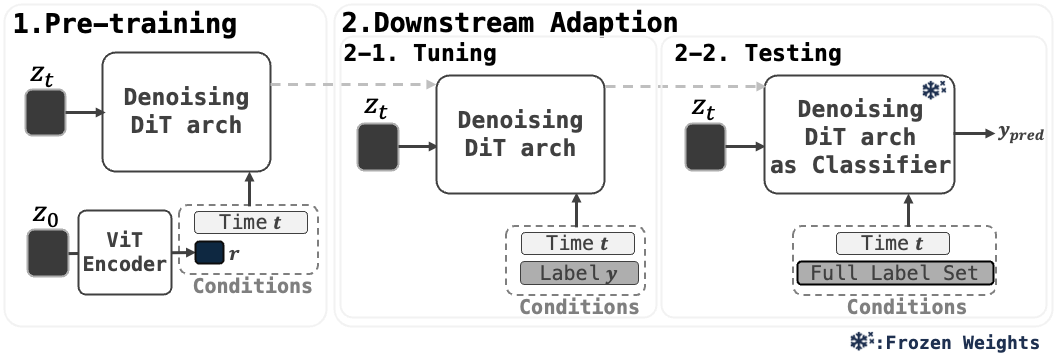}
        \caption{Self-supervised learning using the Representation-Conditioned Latent Diffusion Transformer}
        \label{fig:ssl}
    \end{center}
\end{figure*}

The Representation-Conditioned Latent Diffusion Transformer is a Latent Diffusion Transformer that conditions on representations obtained from ViT. A more detailed architecture is shown in Figure \ref{fig:repcond_dit}. There are two reasons for adopting a mechanism that conditions on representations. First, it enhances the effectiveness of self-supervised learning by more effectively extracting image feature representations within the limited constraint of an unlabeled dataset. Second, we considered that it might be possible to directly use representations that are difficult to obtain explicitly from diffusion models in the downstream task of image classification.

The diffusion process starts from a noise-free latent image $z_0$. The noised latent image $z_t$ at an arbitrary timestep $t$\footnote{$t$ is an integer from $0$ to $T-1$ $(t \in \mathbb{Z}, 0 \leq t < T )$} is represented by $z_t = \gamma_t z_0 + \delta_t \epsilon$. Here, $z_0 = E(x)$ is the latent image obtained by mappingthe image $x$ to the latent space using a VAE encoder $E$\footnote{This is a VAE released by Stable Diffusion, a model trained on ImageNet \shortcite{NIPS2012_c399862d} andfurther fine-tuned on Laion-Humans, LAION-Aesthetics \shortcite{beaumont2021img2dataset}.}, and $\epsilon \sim N(0, I)$ is noise sampled from a Gaussian distribution. Also, $\gamma_t$ and $\delta_t$ are noise scaling factors, satisfying $\gamma_t^2 + \delta_t^2 = 1$. In the Representation-Conditioned Latent Diffusion Transformer, the conditioning representation $r$ is expressed by Equation \eqref{formula:encode}, using a ViT encoder $f_\phi$ with parameters $\phi$.
\begin{equation}
    \label{formula:encode}
    r = f_\phi(z_0)
\end{equation}

The representation $r$ is passed to two locations. First, it is combined with the output of "decode\footnote{Abbreviated as "decode" for simplicity, but it actually includes Norm, Linear, and UnPatchfy layers.}" shown in Figure \ref{fig:repcond_dit} and input into the denoising network. Second, it it combined at  the beginning of each of the N blocks existing within the denoising network. The Represention-Conditioned Latent Diffusion Transformer is trained by predicting the noise $\epsilon$ contained in $z_t$. Specifically, the loss function in Equation \eqref{formula:loss} is minimized.
\begin{equation}
\label{formula:loss}
L(\theta, \phi) = \parallel \epsilon - g_{\theta} (z_t, r, t) \parallel^{2}
\end{equation}
Here, $g_\theta(\cdot)$ is the denoising network (Denoising DiT arch in Figure \ref{fig:repcond_dit}) with parameters $\theta$. The network is trained by being given the representation $r$ dependent on parameters $\phi$ and an arbitary timestep $t$ as conditions, and optimizeing parameters $\theta, \phi$.

While there are several methods for training diffusion models by conditioning on representations \cite{rombach2022highresolution,bordes2022high}, all of them adopt a method of conditioning on representations obtained from a pre-trained encoder. Our method differs in that the ViT encoder that encodes the representation is also trained simultaneously with the training of the denoising network, and the representation is optimized. We expect to acquire representations that promote the optimization of the diffusion model.

\subsection{Self-Supervised Learning with Latent Diffusion Transformer} \label{sec:diff_ssl}
This section describes a new method for using the Latent Diffusion Transformer as a framework for self-supervised learning. This method, which introduces self-supervised learning, aims to solve the problem in Diffusion Classifier, a classification method using diffusion models, where a DiT pre-trained on a large-scale labeled dataset is required. Specifically, by introducing self-supervised learning, it enables the utilization of unlabeled datasets, thereby reducing annotation costs. Furthermore, conventional self-supervised learning approaches like contrastive learning have mitigated the dependence on labeled data by performing representation learning using unlabeled data and applying it to downstream tasks. However, these methods have a challenge in that while they emphasize the global structure of data, they cannot sufficiently perform representation learning that retains local information. In this research, to overcome this challenge, we apply diffusion models to self-supervised learning and aim to acquire more robust and versatile representations.

This method, similar to conventional self-supervised learning approaches, consists of two stages: pre-training on a large unlabeled dataset and a downstream task on a small labeled dataset. Figure \ref{fig:ssl} shows the process of self-supervised learning using the Representation-Conditioned Latent Diffusion Transformer.

\subsubsection{Pre-training}
The Representation-Conditioned Latent Diffusion Transformer is trained using an unlabeled dataset for pre-training. It is trained by minimizing Equation \eqref{formula:loss}. To demonstrate the effect of applying the representation-conditioning mechanism, experiments with an unconditional Latent Diffusion Transformer, which does not have the representation-conditioning mechanism (described later), are also conducted for comparison.

\subsubsection{Downstream Adaption}
\begin{figure}[h]
    \begin{center}
        \includegraphics[scale=0.5]{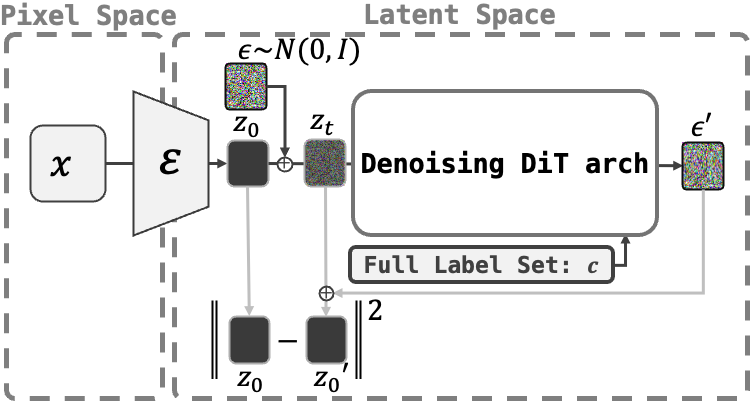}
        \caption{Diffusion Classifier Zero}
        \label{fig:diff_classfiier_zero}
    \end{center}
\end{figure}

In the downstream task of this experiment, we consider image classification. Therefore, we prepare a labeled dataset for the downstream task that does not include pre-training data and conduct experiments. This dataset is broadly divided into a training dataset and a test dataset. The downstream task is divided into two stages: tuning using the training dataset and testing using the test dataset.
\begin{itemize}
    \item \textbf{Tuning} The weights of the denoising network, pre-trained and obtained, are carried over and tuned using the training dataset for the downstream task. Since we are considering an image classification task, the representation-conditioning mechanism is removed from the denoising network, and a class label-conditioning mechanism is added. This is shown in Figure \ref{fig:conditions} (described later).
    \item \textbf{Testing} The weights of the tuned denoising network are frozen, and image classification results are obtained using the test dataset for the downstream task. The detailed approach for obtaining image classification results is described below.
\end{itemize}
Since diffusion models treat internal information as noise, explicit representations do not exist, and we believe it is difficult to consider diffusion models as encoders that output representations. Therefore, we refer to Diffusion Classifier \cite{li2023diffusion} for classification. Diffusion Classifier can be applied to diffusion models trained by conditioning on class labels. For an input $z_t$, it conditions on all labels included in the dataset and predicts the noise $\epsilon$ contained in the input $z_t$. It measures the mean squared error $|| \epsilon - \epsilon' ||^2$ between the predicted noise $\epsilon'$ and the original noise $\epsilon$ for  each class label and obtaines the label that achieves the minimum value among all labels as the predicted label. We achieve classification by evaluating not with the predicted noise $\epsilon'$ but with the predicted latent image $z_0'$. We name this method Diffusion Classifier Zero, and its image is shown in Figure \ref{fig:diff_classfiier_zero}. 
In evaluation by predicted noise, it is difficult to visually distinguish the difference between the original noise and the predicted noise. We propose Diffusion Classifier Zero because we believe it is easier to visually distinguish the difference between the latent image $z_0$ and the predicted latent image $z_0'$. The predicted latent image $z_0'$ is expressed by Equation \eqref{formula:z0prime}, using the trained network $g_{\hat{\theta}}(\cdot)$.
\begin{equation}
    \label{formula:z0prime}
    z_0' = \frac{1}{\gamma_t}(z_t - \delta_t g_{\hat{\theta}}(z_t, y, t))
\end{equation}
Here, since thenetwork after tuing $g_{\hat{\theta}}(\cdot)$ is modeled as a class-conditioned Latent Diffusion Transfromer, the conditions are the class label $y$ and the timestep $t$. Diffusion Classifier Zero obtaines the image classification result as the predicted label $y_{pred}$ by Equation \eqref{formula:diff_classifier}.
\begin{equation}
    \label{formula:diff_classifier}
    y_{pred} = \underset{\mathbf{c}} {\operatorname{argmin}} \large(\mathbb{E}_{t, \epsilon}\left[|| z_0 - \mathbf{z}_0' ||^2\right]\large)
\end{equation}
Here, $\mathbf{c}$ represents the list of all class labels included in the dataset $\mathbf{c} = \{y_i | 0 \leq i < N \}$. Also, $\mathbf{z}_0'$ shown in bold represents the predicted latent image list $\mathbf{z}_0' = \{{z_0'}_i | 0 \leq i < N \}$, and one predicted latent image ${z_0'}_i$ is derived for each class label $y_i$. The mean squared error between the latent image $z_0$ and each predicted latent image ${z_0'}_i$ is measured, and the index number $i$ that showed the minimum value can be obtained as the predicted label $y_{pred}$. Each element ${z_0'}_i$ of $\mathbf{z}_0'$
depends on the hyperparameters $t, \epsilon$. It has been reported \cite{li2023diffusion} that the performance of image classification results is greatly affected, especially by $t$, and the selection of $t$ takes time.

\subsubsection{Conditions for Latent Diffusion Transformer} \label{sec:conditions}
\begin{figure}[h]
    \begin{center}
        \includegraphics[scale=0.55]{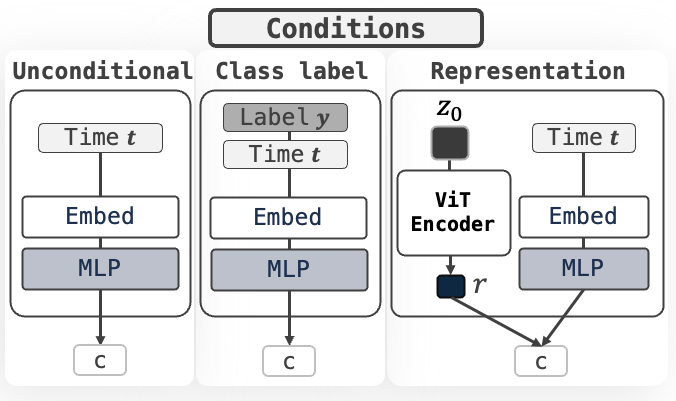}
        \caption{The types of conditions used in this experiment}
        \label{fig:conditions}
    \end{center}
\end{figure}
In this experiment, we adopt the three conditioning mechanisms shown in Figure \ref{fig:conditions}. Unconditional uses timestep embedding. Class-conditioned uses timestep and class label embeddings. Representation-conditioned uses timestep embedding and the representation $r$ obtained from ViT with the latent image $z_0$ as input. We will now explicitly refer to the Latent Diffusion Transformer under each condition as "Unconditional Latent Diffusion Transformer," "Class-Conditioned Latent Diffusion Transformer," and "Representation-Conditioned Latent Diffusion Transformer," respectively.

\section{Experimental Setup} \label{sec:experiment}
Chapter \ref{sec:experiment} describes the detailed settings for the experiments. This experiment focuses on brain CT images in the medical domain. Although the method using the Latent Diffusion Transformer shown in Chapter \ref{sec:propose} can be applied to image classification in general domains, one of the objectives of this research is to create a high-performance classification model with a minimal labeled dataset, considering the reduction of annotation costs. Therefore, we consider using a medical image dataset, which has particularly high annotation costs. Labeling medical images requires expertise and manual work by experienced radiologists or physicians. Furthermore, medical images often have significant imbalances in data volume between classes due to the types of lesions and patient distribution, posing challenges in addressing data imbalance. Tackling the challenge of brain hematoma classification may be applicable not only to analysis in other medical fields such as fundus image analysis and skin lesion diagnosis but also to industrial fields such as semiconductor inspection and manufacturing line anomaly detection, and cultural heritage/art fields such as ancient document and forgery identification, where annotation costs and data imbalance are conceivable.

\subsection{Datasets}
In this experiment, we use a brain CT image dataset. The image data consists of CT scan slices with a size of 512×512 pixels. In this experiment, referencing the preprocessing of prior research \cite{hirano2021classification}, we adjust the contrast based on parameters determined by specialists at each facility, remove artifacts\footnote{Scattering radiation and noise that occur during CT scan imaging or reconstruction.}, and resize the images to 256×256 as image data preprocessing. This dataset is labeled by doctors at each facility for four hematoma markers: "hypodensities," "margin irregularity," "blend sign," and "fluid levels." These hematoma markers are independent and can overlap (it is not a multi-class classification problem). In this experiment, we focus on the presence or absence of hematoma and set annotations such that if any of the four hematoma markers are present, it is a hematoma; if none are present, it is not a hematoma. CT scans were collected from 12 facilities; facilities 2 to 12 are used for pre-training data, and facility 1 is used for downstream task data. Although labels are assigned to the data collected from facilities 2 to 12 for pre-training, in the self-supervised learning by the Latent Diffusion Transformer, it is treated as an unlabeled dataset, and training is performed without using labels. Only in the downstream task are the labels from the data collected from facility 1 used for tuning and testing. 
\begin{table}[h]
\begin{center}
    \caption{Number of data in the dataset and splitting}
    \label{table:dataset}
    \begin{tabular}{c|rrr} \hline\hline
        Phase & Train & Valid & Test \\ \hline
        Pre-training data (2 - 12) & 10,313 & - & - \\ 
        Downsteram task data (1) & 1,424 & 178 & 179 \\ \hline
    \end{tabular}
\end{center}
\end{table}
The number of data points is shown in Table \ref{table:dataset}. The data for the downstream task is further divided into three parts: training data (Train) and validation data (Valid) are used during tuning, and test data (Test) is used during testing.

\subsection{Hyperparameters}
DiT \cite{Peebles2022DiT} offers four model sizes: S, B, L, and XL. The model used in this experiment is the B-size model. The model size is controlled by three factors: embedding dimension, number of blocks, and number of multi-head attention heads\footnote{
Details of the DiT are given in the table below, see \shortcite{Peebles2022DiT}.\\
\begin{tabular}{c|cccc} \hline\hline
    Model & Blocks & Hidden size & Heads & Gflops \\ \hline
    S  & 12 &   384 &  6 &  1.4 \\
    B  & 12 &   768 & 12 &  5.6 \\
    L  & 24 & 1,024 & 16 & 19.7 \\
    XL & 28 & 1,152 & 16 & 29.1 \\ \hline
\end{tabular}
}.
The main hyperparameters for training are shown in Table \ref{table:hyperparam}.
\begin{table}[h]
\begin{center}
    \caption{Main hyperparameter values}
    \label{table:hyperparam}
    \begin{tabular}{c|c} \hline\hline
        Batch size & 16 \\
        Model size & B \\
        Image size & (3, 256, 256) \\
        Latent image size & (4, 32, 32) \\
        Patch size & 2 \\
        Learning rate & 0.0001 \\
        Total time steps: T & 1,000 \\ \hline
    \end{tabular}
\end{center}
\end{table}
Considering the amount of data used, model size, GPU capacity, etc., the batch size is set to 16. The image size in pixel space is (3, 256, 256), and the size in latent space is (4, 32, 32). A larger patch size reduces the number of tokens and thus training cost, but it has been reported that coarser patch division degrades the quality of generated images \cite{Peebles2022DiT}. Therefore, considering training cost and generation quality, it is set to 2. The learning rate and total timesteps are set to 0.0001 and 1,000, respectively, following \cite{Peebles2022DiT}.

The code used in this experiment is implemented in PyTorch \cite{paszke2019pytorch}, and all experiments are trained on a GPU equipped with NVIDIA GeForce RTX 4090 (24GB VRAM) on a Linux OS.

\begin{table*}[h]
\begin{center}
    \caption{Evaluation results by image generation}
    \label{table:eval_gen}
    \begin{tabular}{c|ccccc} \hline\hline
        Models & FID$\downarrow$ & sFID$\downarrow$ & IS$\uparrow$ & Precision$\uparrow$ & Recall$\uparrow$ \\ \hline
        VQ-GAN (unconditional) 
        & 27.69 & 43.10 & 2.981 & \underline{1.00} & \underline{1.00} \\
        VQ-Diffusion (unconditional)
        & 22.55 & 52.95 & 2.795 & 0.98 & 0.54 \\
        \textbf{Class-Conditioned} Latent Diffusion Transformer
        & 25.69 & 21.57 & \underline{3.257} & 0.39 & 0.36 \\ \hline
        \textbf{Unconditional} Latent Diffusion Transformer (ours)
        & 24.42 & 20.98 & 3.117 & 0.40 & 0.39 \\
        \textbf{Representation-Conditioned} Latent Diffusion Transformer (ours)
        & \underline{10.00} & \underline{12.20} & 3.204 & 0.94 & 0.99 \\ \hline
    \end{tabular}
\end{center}
\end{table*}

\subsection{Evaluation Method}
This paper primarily aims to propose a model that achieves high classification performance with a minimal labeled dataset, and while we strive for a configuration focused on classification as much as possible, dealing with a generative model like a diffusion model necessitates evaluation from the perspective of a generative model as well. Therefore, in the experimental results in Chapter \ref{sec:results}, we conduct evaluations from both generation and classification aspects.

\subsubsection{Evaluation of Generation Task}
We evaluate the Representation-Conditioned Latent Diffusion Transformer mainly as a generative model. The models for comparison are the Unconditional Latent Diffusion Transformer, Class-Conditioned Latent Diffusion Transformer, Representation-Conditioned Latent Diffusion Transformer, plus VQ-GAN \cite{esser2021taming} and VQ-Diffusion \cite{gu2021vector}. For VQ-GAN and VQ-Diffusion, we load publicly available pre-trained weights and evaluate them after fine-tuning on brain image data. Both are unconditional, with only timesteps as conditions. For the various Latent Diffusion Transformers, parameters are initialized and trained using pre-training data. Note that while the Unconditional Latent Diffusion Transformer and Representation-Conditioned Latent Diffusion Transformer do not use labels, the Class-Conditioned Latent Diffusion Transformer is trained using labels, albeit from the pre-training data.

For evaluation metrics, we use the Frechet Inception Distance (FID) \cite{heusel2017gans}, a standard metric for evaluating generative models, to measure performance. Conventionally, there are FID-50k measured from 50,000 samples and FID-10k from 10,000 samples. We generate 10,000 samples, comparable to the scale of our training set, and measure and report FID-10k. FID is known to be sensitive to minor differences in implementation \cite{parmar2022aliased}. Therefore, to ensure accurate comparison, all FID values reported in this paper are calculated by converting the generated samples into npz format binary files and using ADM's TensorFlow evaluation \cite{dhariwal2021diffusion}. Furthermore, Inception Score (IS) \cite{salimans2016improved}, Spatial FID (sFID) \cite{nash2021generating}, and Precision/Recall \cite{kynkaanniemi2019improved} are reported as supplementary evaluation metrics.

\subsubsection{Evaluation of Classification Task}
Similar to the evaluation of the generation task, we evaluate the Representation-Conditioned Latent Diffusion Transformer mainly as a classification model. The models for comparison are broadly categorized by the presence or absence of pre-training. First, as baselines without pre-training, we report the results of ResNet \cite{he2016deep} and ViT \cite{dosovitskiy2020image}. These are trained on the downstream task's training data without using the pre-training dataset and tested on the test data. Next, as models with pre-training, we report the results of the Unconditional Latent Diffusion Transformer, Representation-Conditioned Latent Diffusion Transformer, plus DINOv2 \cite{oquab2023dinov2}, a powerful contrastive learning method. All parameters are initialized, pre-trained without labels on the pre-training dataset, and then tuned on the downstream task's training data. Finally, they are tested on the downstream task's test data, and the results are shown. The reason for not using the Class-Conditioned Latent Diffusion Transformer is that since the pre-training conditions are set to unconditional and representation-conditioned, if training with class labels were performed during pre-training, the Class-Conditioned Latent Diffusion Transformer would have trained on class label information for the amount of pre-training data, which is considered inappropriate for comparison. Therefore, the Class-Conditioned Latent Diffusion Transformer is excluded.

For evaluation metrics, performance is evaluated by Accuracy, a standard metric for evaluating classification models. Since Accuracy alone can make proper evaluation difficult if the class distribution of data is imbalanced, F1-Score, Recall, and Precision are used as supplementary metrics.

\begin{table*}[h]
\begin{center}
    \caption{Evaluation results by hematoma classification}
    \label{table:eval_classify}
    \scalebox{1.}{
    \begin{tabular}{c|c|rrrr} \hline\hline
        Pre-trained & Models & Accuracy & F1-score & Recall & Precision \\ \hline
        \ding{55} & ResNet 50 & .7150 & .4950 & \underline{.8928} & .3424 \\
        \ding{55} & ViT B     & .8659 & .5200 & .4642 & .5909 \\
        \ding{51} & DINOv2    & .8882 & .6875 & .7857 & .6111 \\ \hline
        \ding{51} & \textbf{Unconditional} (ours)
                  & .8882 & .7058 & .8571 & .6000 \\
        \ding{51} & \textbf{Representation-Conditioned} (ours)
                  & \underline{.9497} & \underline{.8235} & .7500 & \underline{.9130} \\
        \hline
    \end{tabular}
    }
\end{center}
\end{table*}

\section{Results} \label{sec:results}
Chapter \ref{sec:results} shows the experimental results and discusses them. Section \ref{sec:result_gen} shows the results related to generation, and Section \ref{sec:result_classify} shows the results related to classification.

\subsection{Generation Results} \label{sec:result_gen}
Table \ref{table:eval_gen} shows the evaluation of generative models. Regarding FID and SFID, the Representation-Conditioned Latent Diffusion Transformer achieved the smallest values compared to other models. We believe this result was obtained because providing representations containing information from the original image enables the generation of images closely resembling the original, leading to a closer data distribution.

\begin{figure}[h]
    \begin{center}
        \includegraphics[scale=0.70]{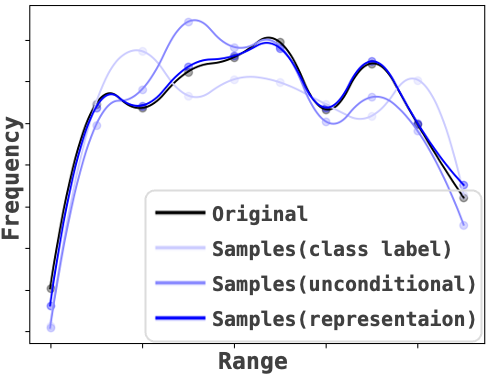}
        \caption{Distribution of pre-training data (black) and distribution of samples generated under different conditions (blue)}
        \label{fig:distribution}
    \end{center}
\end{figure}

Figure \ref{fig:distribution} shows the distribution of image data. The horizontal axis represents the range where data is distributed (discrete values), and the vertical axis represents the number of data points existing in that range. Feature representations (2048 dimensions) output from Inception-v3 \cite{szegedy2015rethinking} were compressed to 1 dimension using t-SNE \cite{vanDerMaaten2008}, a data compression model, rounded to discrete values, and their counts were plotted. Black represents the pre-training data distribution, and the three shades of blue represent the data distribution of generated images. Light blue is for class-conditioned generated samples, pale blue for unconditional generated samples, and blue for representation-conditioned generated samples. Supporting the FID values in Table \ref{table:eval_gen}, it can be seen that the data distribution of samples generated by the Representation-Conditioned Latent Diffusion Transformer is most similar to the pre-training data distribution. None of the Latent Diffusion Transformers under each condition show signs of mode collapse\footnote{A problem encountered especially in GAN training, where many of the generated images include duplicates (modes) and lack variety.}, and they achieve the generation of diverse brain images.

\begin{figure}[h]
    \begin{center}
        \includegraphics[scale=0.35]{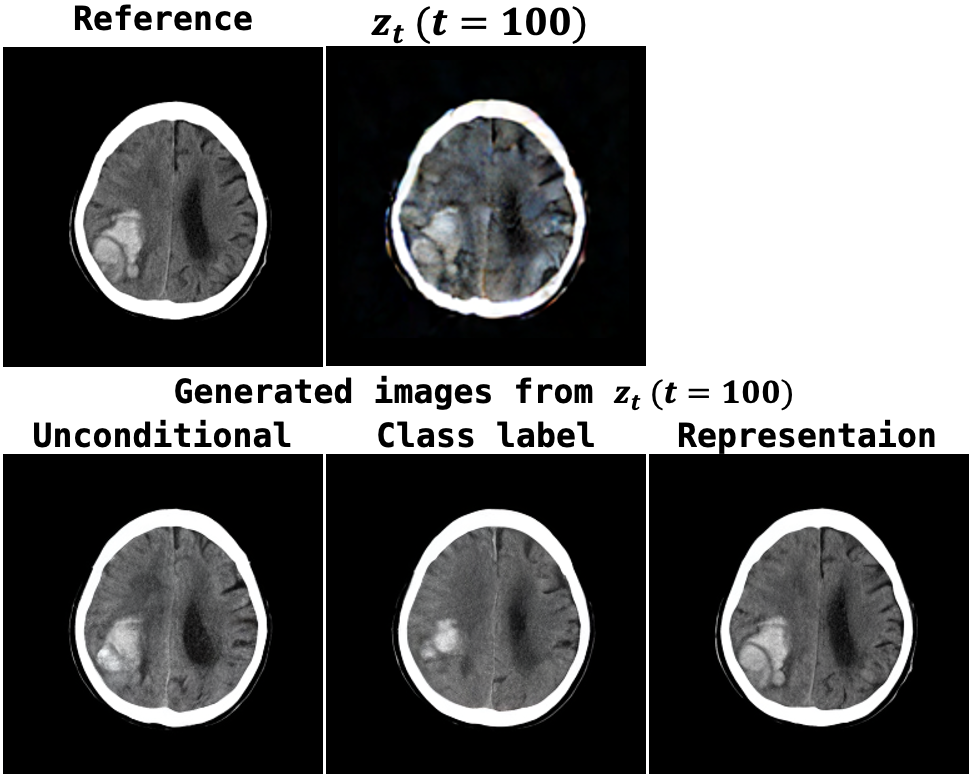}
        \caption{Quality of Images Generated by our Latent Diffusion Transformers}
        \label{fig:samples}
    \end{center}
\end{figure}

To evaluate the quality of images generated by various Latent Diffusion Transformers, Figure \ref{fig:samples} is shown. Originally, the image generation process of diffusion models involves $T$ steps of denoising from complete noise $\epsilon \sim N(0, I$, but it is difficult to generate consistently controlled images for comparison with models under different conditions. Therefore, instead of starting from the timestep $t=T(=1000)$ where complete noise is obtained, we input the latent image $z_t (=z_{100})$ at an intermediate timestep $t=100$, and compare the images generated after $t(=100)$ steps of denoising. Figure \ref{fig:samples} shows the original image (top left), the image of $z_t$ decoded into pixel space (top right), and generated samples for each condition (bottom). First, regarding the hematoma, which appears as white-gray within the skull, its shape differs for each condition. In terms of reproducibility of the original image, the representation-conditioned one is the most reproducible, followed by unconditional, and lastly class-conditioned. The class-conditioned example shows the lower half of the hematoma missing, and its shape is significantly different compared to the original image. Next, regarding the lateral ventricles shown in black within the skull, the representation-conditioned sample is also the most reproducible. Finally, focusing on the wrinkles in the lower right of the intracranial area, only the representation-conditioned sample can reproduce them, while the unconditionally and class-conditioned generated samples either have the wrinkles missing or only show faint black marks. From the above, it can be seen that in terms of reproducing the original image, the representation-conditioned generated sample is of the highest quality.

\subsection{Classification Results} \label{sec:result_classify}
Table \ref{table:eval_classify} shows the hematoma classification results as an evaluation from the classification perspective. These results were obtained using the test data for the downstream task. Our proposed Representation-Conditioned Latent Diffusion Transformer shows the highest accuracy, achieving an improvement of +6.15\% in Accuracy and +13.60\% in F1 Score compared to DINOv2. The next best results were from the Unconditional Latent Diffusion Transformer, which had the same Accuracy as DINOv2 but achieved a +1.83\% improvement in F1 Score. From the above, it was shown that in this experimental setting, our proposed method using self-supervised learning with Latent Diffusion Transformers achieves higher performance than DINOv2, a powerful method even among contrastive learning techniques.

Next, we discuss the differences arising from the presence or absence of the representation-conditioning mechanism in the pre-training stage. We investigate this difference by comparing the unconditional and representation-conditioned versions. Table \ref{table:eval_classify} shows a difference of 6.15\% in Accuracy and 11.77\% in F1 Score. We consider the reasons why incorporating the representation-conditioning mechanism achieves this significant performance improvement.

\begin{figure}[h]
    \begin{center}
        \includegraphics[scale=0.48]{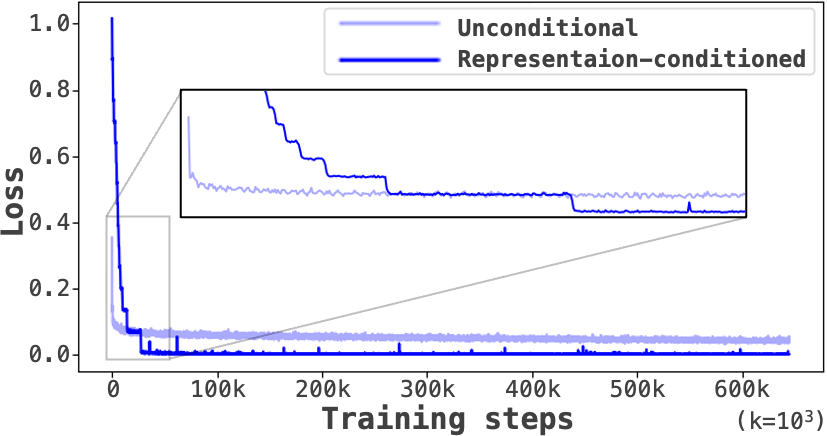}
        \caption{Loss transition during pre-training}
        \label{fig:loss}
    \end{center}
\end{figure}

First, we compare the losses during the pre-training stage. Figure \ref{fig:loss} shows the transition of loss for each model. Light blue indicates the loss for the Unconditional Latent Diffusion Transformer, and blue indicates the loss for the Representation-Conditioned Latent Diffusion Transformer. It is clear that the Representation-Conditioned Latent Diffusion Transformer succeeds in reducing the loss to a smaller value. Comparing the minimum loss values for each model, the unconditional model achieves $3.13 \times 10^{-2}$, while the representation-conditioned model achieves $3.39 \times 10^{-4}$, indicating a difference of approximately $10^2$ scale. The loss discussed here is the mean squared error between the noise and the predicted noise, as shown in Equation \eqref{formula:loss}. It is thought that incorporating the representation-conditioning mechanism improves the accuracy of the predicted noise.

\begin{figure}[h]
    \begin{center}
        \includegraphics[scale=0.5]{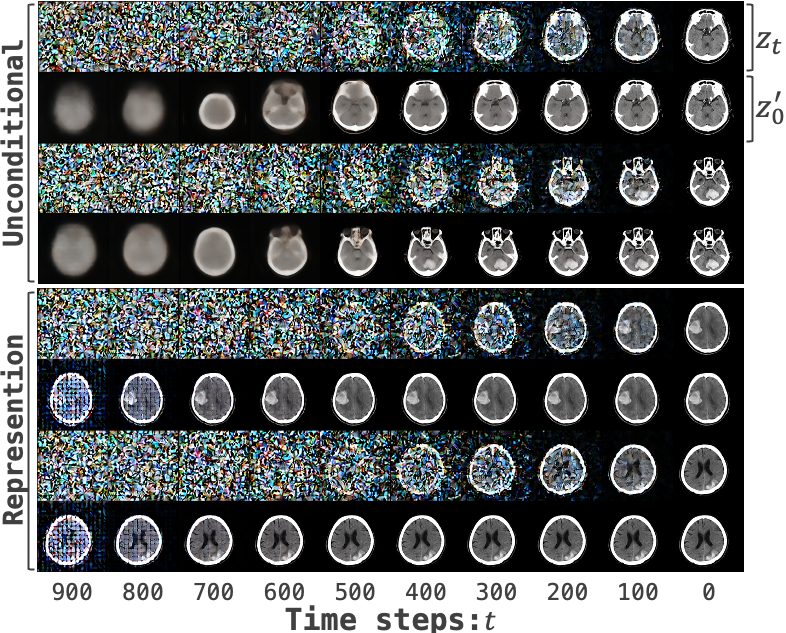}
        \caption{$z_t$ and $z_0'$ according to time step $t$}
        \label{fig:comparison_denoising}
    \end{center}
\end{figure}

Next, we compare the predicted latent images. The predicted latent image $z_0'$ is directly used in Diffusion Classifier Zero, and its accuracy depends on the timestep $t$. Figure \ref{fig:comparison_denoising} shows the noised latent image $z_t$ (upper row) and the predicted latent image $z_0'$ (lower row) for timestep $t$. Two samples each for unconditional and representation-conditioned are shown; the top two are unconditional results, and the bottom two are representation-conditioned results. Comparing unconditional and representation-conditioned, it is clear that the quality of the predicted latent image $z_0'$ differs significantly. $z_0$ at $t=0$ (far right) can be considered the original image. However, the timing at which the prediction of the original image becomes possible differs between unconditional and representation-conditioned. For the unconditional case, an image similar to the original image is predicted around $t=400$. For $z_0'$ at $t\geq600$, the image is significantly different from the original image $z_0$and is a low-quality, blurred image. On the other hand, for the representation-conditioned case, it seems that the original image is almost predicted at around $t=800$. Furthermore, at $t=900$, an image similar to the original is already predicted, overlapping with mesh-like noise. For $t\leq700$, as the timestep decreases, the prediction transitions to become closer to the original image in finer details. From this, it is considered that the Representation-Conditioned Latent Diffusion Transformer can accurately predict the latent image $z_0'$ at an earlier (larger) timestep compared to the Unconditional Latent Diffusion Transformer, and as the timestep decreases, it performs finer adjustments to the prediction accuracy.

The above discussion covered comparisons of loss and predicted latent images $z_0'$. From the loss comparison, it was found that incorporating the representation-conditioning mechanism enables deeper training on a $10^2$ scale. From the comparison of predicted latent images $z_0'$, it was found that incorporating the representation-conditioning mechanism allows the original image to be predicted at an earlier stage, and fine-grained adjustment of prediction accuracy can be performed over a wide timestep space. These two points are thought to contribute to the acquisition of representations effective for image classification.

\section{Conclusion} \label{sec:conclusion}
In this research, we proposed a framework that utilizes diffusion models as pre-training for self-supervised learning and applies them to subsequent image classification tasks. We also newly proposed a Representation-Conditioned Latent Diffusion Transformer. Regarding image generation, the Representation-Conditioned Latent Diffusion Transformer showed higher performance compared to other conditions, demonstrating that the representation-conditioning mechanism improves the quality of generated images. Regarding image classification, the Representation-Conditioned Latent Diffusion Transformer achieved an improvement of +6.15\% in Accuracy and +13.60\% in F1 Score compared to DINOv2, realizing performance improvement through an approach different from conventional contrastive learning. Furthermore, comparison with the Unconditional Latent Diffusion Transformer suggests that conditioning on representations facilitates training and enables finer predictions, thereby achieving the acquisition of representations effective for image classification.

\bibliography{aaai25}
\end{document}